\pdfoutput=1

\documentclass[11pt]{article}

\usepackage[final]{acl}

\usepackage{times}
\usepackage{latexsym}
\usepackage{amsmath}
\usepackage{amssymb}
\usepackage{booktabs}
\usepackage{tcolorbox}
\usepackage{bbding}
\usepackage[T1]{fontenc}

\usepackage[utf8]{inputenc}

\usepackage{microtype}
\usepackage{inconsolata}
\usepackage{colortbl}
\usepackage{graphicx}
\usepackage{multirow}

\title{LongDPO: Unlock Better Long-form Generation Abilities for LLMs via Critique-augmented Stepwise Information}

\author{
 \textbf{Bowen Ping\textsuperscript{1}},
 \textbf{Jiali Zeng\textsuperscript{2}},
 \textbf{Fandong Meng\textsuperscript{2}},
 \textbf{Shuo Wang\textsuperscript{3}},
\\
 \textbf{Jie Zhou\textsuperscript{2}},
 \textbf{Shanghang Zhang\textsuperscript{1}}\Envelope
\\
\\
 \textsuperscript{1}State Key Laboratory of Multimedia Information Processing, \\ School of Computer Science, Peking University, \\
 \textsuperscript{2}Pattern Recongnition Center, WechatAI, Tencent Inc, \\
 \textsuperscript{3}Dept. of Comp. Sci. \& Tech., Tsinghua University, Beijing, China
\\
 \small{
   \textbf{Correspondence:Shanghang Zhang} \href{sh}{shanghang@pku.edu.cn}
 }
}

\begin{document}
\maketitle
\begin{abstract}

Recent advancements in large language models (LLMs) have markedly improved their capacity to handle long text inputs; however, current models, including GPT-4o, still exhibit unsatisfactory performance in long-form generation. Generating high-quality long-form content still remains a significant challenge. In this paper, we present LongDPO, a novel approach designed to enhance long-form text generation through step-level supervision. By leveraging Monte Carlo Tree Search (MCTS) to collect stepwise preference pairs and employing a global memory pool to maintain factual accuracy, LongDPO effectively mitigates issues such as inconsistencies that are prevalent in long-context LLMs. Furthermore, we integrate critique-augmented generation to refine the selected preference pairs. Following the collection of stepwise preference pairs, we apply stepwise preference learning for fine-grained optimization. Experimental results demonstrate that our method enhances performance on long-form generation benchmarks (e.g.~LongBench-Write) while maintaining nearly lossless performance on several general benchmarks. \footnote{\ \ Code and models will be publicly available at \url{https://github.com/pingbowen23/LongDPO}.}
\end{abstract}

\begin{figure}[ht]
  \centering
  \includegraphics[width=0.99\linewidth]{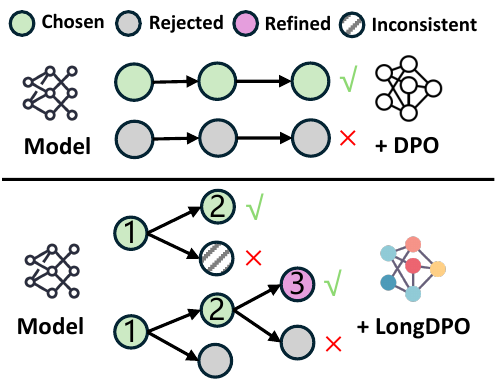}
  \caption{
The above refers to outcome supervision, which directly provides feedback for extended sequences in long-form generation tasks. Below is LongDPO uses process supervision with a global memory to maintain factual consistency, and external critiques to refine low-reward chosen candidates.
  }
  \label{fig1}
\end{figure}

\section{Introduction}

Recent advancements in large language models (LLMs)~\citep{zhou2024llmtimesmapreducesimplifiedlongsequenceprocessing,xiao2024duoattentionefficientlongcontextllm,xiao2024infllmtrainingfreelongcontextextrapolation,wang-etal-2024-lora-flow,ping2024deltacometrainingfreedeltacompressionmixedprecision}, have significantly enhanced their capacity to process long text sequences with models like GPT-4o now capable of handling contexts up to 128K tokens~\citep{openai2024gpt4ocard,yang2025qwen3technicalreport}. 
Despite these strides, there has been less emphasis on the models' ability to generate better long-form text outputs.
The capability to produce long-form content is essential for various real-world applications, including writing academic papers, novels, and scripts in literature, generating legal contracts in law, and producing repository-level code in technology~\citep{bai2024longwriterunleashing10000word,wang2024autosurveylargelanguagemodels}. 
However, many LLMs still struggle to generate content exceeding 2,000 words~\citep{DBLP:conf/emnlp/PhamSI24,bai2024longwriterunleashing10000word}, highlighting the need for further advancements in this area.

Previous research has explored methods to extend the output window by creating long-form training data and leveraging preference learning. For example, Suri~\citep{DBLP:conf/emnlp/PhamSI24} creates various instructions for the same response and performs outcome-level preference optimization. LongWriter~\citep{bai2024longwriterunleashing10000word} employs an agent-based pipeline that decomposes ultra-long generation tasks into subtasks to build a long-form dataset, followed by supervised fine-tuning and DPO. These approaches primarily rely on outcome supervision~\citep{DBLP:conf/iclr/LightmanKBEBLLS24} during DPO, which provides feedback on the final result, for long-form generation tasks.


Nevertheless, long-context LLMs are more prone to produce responses with issues such as logical inconsistencies, fabricated content, and failure to fully meet query requirements~\citep{zhang2024longrewardimprovinglongcontextlarge}. 
These challenges make outcome supervision, which directly provides feedback for a long sequence, particularly problematic. In contrast, process supervision involves supervising each intermediate step, which offers more granular and precise feedback. Furthermore, process supervision specifies the exact location of low-quality steps, thereby facilitating the refinement of these steps~\citep{DBLP:conf/iclr/LightmanKBEBLLS24}. Consequently, breaking down a long sequence into intermediate steps and supervising these shorter steps could be a more effective strategy.



In this paper, we introduce LongDPO, which enhances long-form generation capabilities through step-level supervision. LongDPO first constructs preference data with stepwise supervision and then performs stepwise learning. Specifically, we use Monte Carlo Tree Search (MCTS)~\cite{6145622} to collect stepwise preference pairs. 
Considering that long-context LLMs are prone to generating inconsistent content, leading hallucinations~\citep{zhang2024longrewardimprovinglongcontextlarge}, we incorporate a global memory pool to improve the factual consistency of the selected preference pairs. 
Additionally, the quality of candidates generated heavily relies on the original model's inherent capability. Simply searching for candidates is both inefficient and ineffective~\citep{qi2024mutualreasoningmakessmaller}. To address this, we propose critique-augmented generation to obtain better candidates for the selected preference pairs.



After gathering the stepwise preference pairs, we propose employing a stepwise DPO for fine-grained learning. 
As illustrated in Figure~\ref{fig1}, traditional DPO applies sample-wise supervision directly, which can lead to a less pronounced reward margin, complicating the learning process~\citep{lai2024stepdpostepwisepreferenceoptimization}. 
In contrast, LongDPO utilizes fine-grained learning at each step, which has the potential to produce superior results.

We evaluate long-form generation capabilities using LongBench-Write-en and LongGenBench~\citep{bai2024longwriterunleashing10000word, wu2024spinning}, which assess text generation length, quality, and adherence to instructions. 
Additionally, we use general benchmarks such as TruthfulQA~\citep{DBLP:conf/acl/LinHE22} to measure overall task performance. Our method, built on Llama- and Qwen-based backbones, outperforms their vanilla DPO versions in long-form generation tasks while maintaining near-lossless performance on general tasks.



Our contributions can be summarized as follows:
\begin{itemize}
    \item We introduce LongDPO, which facilitates step-wise, fine-grained learning for long-form text generation. 
    \item We employ MCTS to create step-level preference data, incorporating a memory pool to enhance factual consistency and external critiques to gather higher-quality preference pairs for long-form generation.
    \item The experimental results and in-depth analysis demonstrate the effectiveness of our method in long-form generation tasks. 
\end{itemize}

\begin{figure*}[t]
  \centering
  \includegraphics[width=0.98\linewidth]{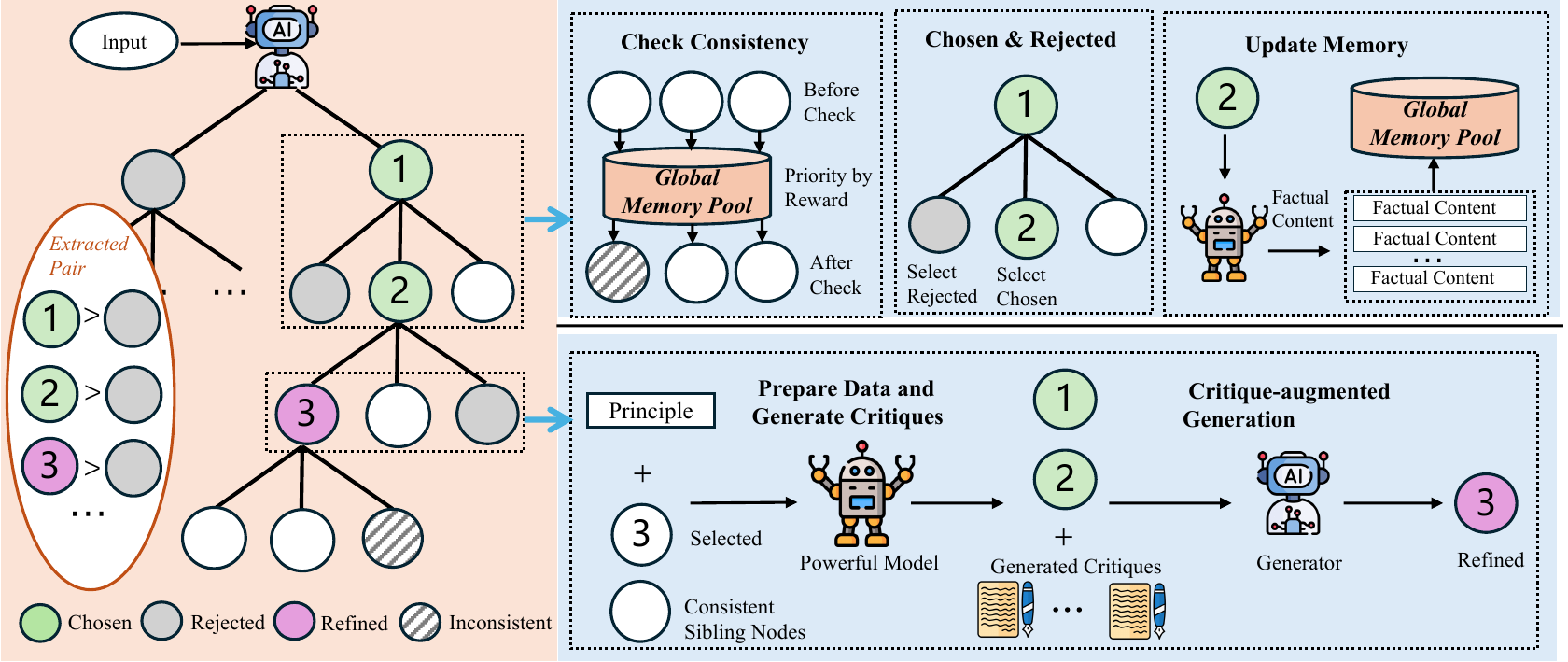}
  \caption{
The pipeline of LongDPO. LongDPO incorporates process supervision and MCTS to collect stepwise preference data. During the selection phase, LongDPO uses the global memory pool to filter out candidates that may result in inconsistency, then selects the highest-scoring one as the chosen candidate, with another randomly selected as the rejected candidate. During tree expansion, LongDPO leverages external critiques only for low-reward chosen candidates. Then the collected preference pairs are used for step-level DPO training. 
  }
  \label{fig2}
\end{figure*}
\section{Related Work}
\textbf{Long Context LLMs} Some studies explore to extend the input context window, using training-based methods like~\citep{bai-etal-2024-longalign,munkhdalai2024leavecontextbehindefficient,DBLP:conf/icml/FuPNYHK024} and training-free methods, such as~\citep{DBLP:conf/iclr/PengQFS24,DBLP:conf/iclr/XiaoTCHL24,DBLP:conf/icml/DingZZXSX0Y24}. Many LLMs can support input context windows of 128K. However, far fewer are capable of generating outputs exceeding 2K words in length. Recent studies~\citep{DBLP:conf/emnlp/PhamSI24,bai2024longwriterunleashing10000word} have employed outcome supervision to extend the output window. Most recently, ~\citet{zhang2024longrewardimprovinglongcontextlarge} proposed LongReward, which is orthogonal to our work. However, in addition to the instruction and response, it requires an additional reference long document as input, which limits its applicability in both outcome and process supervision. Another line of exploration in long-text generation, such as hierarchical writing and recurrent prompting~\citep{quan2024languagemodelsselflengthengenerate,xi2025omnithinkexpandingknowledgeboundaries,wang2024weaverfoundationmodelscreative}, is orthogonal to our method.

\textbf{Process Supervision in Preference Learning} 
Recently, scaling inference-time compute has become increasingly popular~\citep{chen2024optimaoptimizingeffectivenessefficiency,setlur2024rewardingprogressscalingautomated,snell2024scalingllmtesttimecompute}. Process supervision with MCTS can further enhance models' reasoning abilities~\citep{tian2024selfimprovementllmsimaginationsearching,zhang2024chainpreferenceoptimizationimproving,zhang2024accessinggpt4levelmathematical}. Recent studies~\citep{wang2024selfimprovementllmsmctsleveraging,xu2024sramctsselfdrivenreasoningaugmentation} use MCTS in both math and code tasks. In addition to MCTS, ~\citet{zhao2024marcoo1openreasoningmodels} also incorporate self-reflection. ~\citet{cheng2024sparselfplaytreesearchrefinement} employ tree search and train a refiner for iterative optimization. In this work, we primarily focus on exploring the potential of process supervision with MCTS in long-form generation.

\textbf{Use LLM to Critic} The LLM-generated critiques are able to provide additional information and have been widely applied~\citep{madaan2023selfrefineiterativerefinementselffeedback,yuan-etal-2024-llmcrit}. CriticGPT~\citep{mcaleese2024llm}, trained using reinforcement learning, can generate critiques that surpass those produced by humans. Recent studies~\citep{ankner2024critiqueoutloudrewardmodels, ye2024improvingrewardmodelssynthetic} use self-generated critiques for each piece of preference data, which are used to train reward models.~\citet{yu2024selfgeneratedcritiquesboostreward} further uses an instance-level critiques filter to reduce conflicts.

\section{LongDPO} Our method consists of two main parts: 1) collecting stepwise preference data, and 2) using the collected preference data for DPO training.
\subsection{Stepwise Preference Data Construction}
Currently, MCTS has demonstrated its potential in reasoning tasks which employs an additional reward model to better preference data at each reasoning step~\citep{chen2024alphamathzeroprocesssupervision,xie2024montecarlotreesearch}, enabling 7B models to achieve performance comparable to GPT-o1~\citep{guan2025rstarmathsmallllmsmaster}. Intuitively, long-form generation may also be learned by collecting stepwise preference data. We will elaborate on collecting preference data in the following.

\subsubsection{Overview} MCTS executes four procedures: selection, expansion, evaluation, and back-propagation. To be specific, our tree is executed according to the following:
    \begin{itemize}
        \item \textbf{Selection}: We select the node to be expanded using Equation~\ref{ucb} with a global memory pool to filter out inconsistent nodes.
        \begin{equation}
        \mathrm{UCB_{i}} = \alpha \times \sqrt{2 \times \ln \left( \frac{N_i}{1 + n_i} \right)} + v_{i},
        \label{ucb}
    \end{equation}
    where \( n_{i} \) and \( N_{i} \) represent the visit count and the parent visit count of the node, respectively. \( \alpha \) is a scalar that balances exploration and exploitation. \( v_{i} \) denotes the value of the node, and we use the average reward provided by a reward model.
    \item \textbf{Expansion}: For each node to be expanded, we generate several child nodes using a sampling-based algorithm \citep{holtzmancurious}. 
    \item \textbf{Evaluation}: In terms of evaluating each node, we assess each node using the value provided by a reward model, as previous work has demonstrated its effectiveness~\citep{ wang2024selfimprovementllmsmctsleveraging,wang2024litesearchefficacioustreesearch}. We consider seven principles to evaluate each node. Each principle is rated between 1 and 5, as detailed in Appendix~\ref{text_reward}.
    \item \textbf{Back-propagation}: We update the parent node using the value of the leaf nodes and also update the parent node's visit count.
\end{itemize}

Specifically, given a query \( q \), during the expansion phase, the node in layer \( t \) is represented as \( s_t \). The newly node $s_{t+1}$ is generated using the Equation~\ref{step}: 
\begin{equation}
    \label{step}
    s_{t+1} = \pi_{\theta}(q \oplus s_{1} \oplus s_{2}\oplus \dots \oplus s_{t}),
\end{equation}
where $\pi_{\theta}$ is the generator, and $\oplus$ represents the concatenation operation. In each evaluation phase, its corresponding value is evaluated as:
\begin{equation}
    \label{reward}
    r_{s_{t+1}} = \Theta(q \oplus s_1 \oplus s_2 \oplus \dots \oplus s_t, s_{t+1}),
\end{equation}
where \( r_{s_{t+1}} \) is the average reward of the seven principles, \( \Theta \) is the reward model used to evaluate the reward of \( s_{t+1} \) as the suffix. When reaching each leaf node, the back-propagation phase is executed. At each selection phase, we use Equation~\ref{ucb} along with a global memory pool to make selections, as detailed in the next subsection.

\subsubsection{Preference Pair Extraction}
We use a global memory pool $M$ storing relevant factual context $\{m_1,m_2,\dots,m_k \}$ to check consistency before selection. Specifically, after the expansion phase, we visit the nodes in descending order of their UCB scores in Equation~\ref{ucb}. We break the currently visited node $s_{cur}$ into contexts of 128 words, resulting in \(\{ s_{cur_1}, s_{cur_2}, \dots, s_{cur_j} \} \), each $s_{cur_j}$ has 128 words, and calculate the similarity score using each $m_k$ in $M_t$ as a query. 
\begin{equation}
    \label{similarity_score}
     \mathrm{{sim}_{kj}} = E(m_{k}) \times E(s_{cur_j})^T,
\end{equation}
where  $\rm sim_{kj}$ is the similarity score, $E(x)$ represents get the embedding of $x$, we use gte-Qwen2-1.5B-instruct\footnote{\url{https://huggingface.co/Alibaba-NLP/gte-Qwen2-1.5B-instruct}} as embedding model.
Then, we use the similarity score to filter irrelevant context for each $m_k$.
\begin{equation}
    \label{support_fact}
    A_{k} = \{ s_{cur_j} \mid \rm sim_{kj} \geq \delta \},
\end{equation}
where $\delta$ the similarity threshold is set to 0.8. Finally, we use each \( m_k \) and its corresponding supported context $A_{k}$ to check for any inconsistencies using model $\Theta$ using templates in Appendix~\ref{judge_fact}. Finally, if no inconsistencies are found, we select $s_{cur}$ for the next expansion phase. Otherwise, we will visit the next candidate node without expanding the current one further.


After finishing each selection phase, the memory pool \( M\) is also updated accordingly. To be specific, after selecting the node $s_t$, we extract the factual content of $s_t$ using the model $\Theta$ and employ $\Theta$ to verify the extracted factual content to ensure that they are factually correct as much as possible using templates in Appendix~\ref{find_fact}. We retain only the factual content $\{m_1,m_2,\dots,m_{k'} \}$ that does not conflict with the internal knowledge of \( \Theta \). Then, we update the memory correspondingly $M_t = M_{t-1} \cup \{m_1,m_2,\dots,m_{k'} \}$. 

If memory \( M \) is empty, we skip the consistency check and proceed directly to the selection phase and update the memory. When we select \( s_t \), we only use the factual content stored in \( M_{t-1} \), which contains the factual content from the first layer up to the \( t-1 \) layer.

For each layer of the tree, we select one pair for preference learning: the node with the highest average reward and no consistency errors is selected as the chosen candidate $s_{win}$, while another node is randomly selected as the rejected candidate $s_{lose}$.



\subsection{Chosen Candidates Refinement using Critiques}
After collecting preference pairs for long-form generation, we then randomly select 1,000 pairs and only analyze the average reward of the chosen candidate in each pair, as shown in Figure~\ref{reward_distribution}. On the one hand, many of the chosen candidates in each preference pair have low rewards which may lead to suboptimal performance. On the other hand, the large reward discrepancies between different samples could result in unstable training~\citep{wu2024betadpodirectpreferenceoptimization}.

One way to improve performance is by expanding the search space. On the one hand, this is inefficient, especially in the context of long-form generation. On the other hand, recent studies~\citep{brown2024largelanguagemonkeysscaling,qi2024mutualreasoningmakessmaller} have shown that the gains from this approach are limited. Therefore, we propose leveraging external critiques to guide the generator in text generation, as self-critique relies on the model's inherent capabilities. Recent studies have highlighted its instability in driving improvement~\citep{qi2024mutualreasoningmakessmaller,zhang2024understandingdarkllmsintrinsic}.

To be specific, we collect the chosen candidates in each preference pair with average rewards below the threshold \( \eta \) for refinement, as shown in Equation~\ref{thresh_hold}.

\begin{equation}
    \label{thresh_hold}
    S_{R} = \{ s_{win} \mid r_{s_{win}} \leq \eta \},
\end{equation}
where $s_{win}$ and $r_{s_{win}}$ represent the chosen candidate of the collected preference pair and the corresponding average reward. We only refine the chosen candidates, set \( \eta = 2.5 \), and have conducted an ablation study. 

\textbf{Collect Data for Critiques Generation} $S_{R}$ contains the chosen candidates that need to be refined. Next, we prepare the data for the generation of critiques. Specifically, each data is a triplet \( (\text{principle}_u, s_{sib}, s_{win}) \), where \( \text{principle}_u \) is used in the evaluation phase in MCTS to assess the reward of each node, \( s_{win} \) is the chosen candidate to be refined, and \( s_{sib} \) is the sibling node of \( s_{win} \), which serves as an example of refinement as illustrated in Figure~\ref{fig2}. Detailed principles are given in Appendix~\ref{text_reward}.

We construct each pair as the following: for each \( \text{principle}_u \) and \( s_{win} \), if there exists a \( s_{sib} \) whose reward is greater than \( s_{win} \) under \( \text{principle}_u \), the tuple \( (\text{principle}_u, s_{sib}, s_{win}) \) forms a pair to generate critiques.

\begin{figure}[ht]
  \centering
  \includegraphics[width=0.62\linewidth]{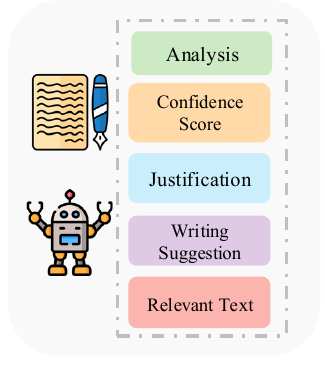}
  \caption{Main body of generated critiques which have detailed in Appedix~\ref{refine_template}
  }
  \label{refine_main_body}
\end{figure}

\textbf{Generate critiques} Next, we use the reward model $\Theta$ to generate critiques for each triplet using template in Appendix~\ref{refine_template}. Figure~\ref{refine_main_body} has shown the main body of the critiques. ``Analysis,'' ``Justification,'' and ``Relevant Text'' are used to enhance the accuracy of the analysis, while the ``Confidence Score'' helps assess the model's confidence in the accuracy of its analysis. ``Writing Suggestion'' provides recommendations for improvement.

\textbf{Critique-augmented Generation} For each \( s_{win} \), we utilize its corresponding critiques \( \{z_1, z_2, \dots, z_\lambda\} \), sorted in descending order by ``Confidence Score,'' to perform critique-augmented generation. Specifically, if \( s_{win} \) is selected in layer \( t+1 \), we rewrite Equation~\ref{step} as follows:
\begin{equation}
    \label{critic-augmented}
        s_{win\_new} = \pi_{\theta}(q \oplus s_{1} \oplus \dots \oplus s_{t} \oplus  \dots \oplus z_\lambda\ ),
\end{equation}
where we use each ``Writing Suggestion'' from \( z_\lambda\ \), with a maximum of three. Then, we use the refined data for DPO training.

\subsection{LongDPO Training Objective}
Previous work on outcome supervision in long-form generation directly utilizes the complete chosen and rejected responses for training~\citep{DBLP:conf/emnlp/PhamSI24,bai2024longwriterunleashing10000word}.
\begin{multline} \label{eq:dpo}
    \mathcal{L}_{DPO} = -\mathbb{E}_{(q, y_w, y_l) \sim D} \Big[ \log \sigma \big( \\ \beta \log\frac{\pi_\theta(y_w|q)}{\pi_{ref}(y_w|q)}
    - \beta \log\frac{\pi_\theta(y_l|q)}{\pi_{ref}(y_l|q)} \big) \Big],
\end{multline}
where $y_{w}$ and $y_{l}$ is the chosen and rejected response, respectively and $\pi_{ref}$ is the reference model. $D$ is the pair-wise preference dataset, $\sigma$ is the sigmoid function, and $\beta$ controls the degree of deviation from the reference model.

In LongDPO, the response \( y \) is decomposed into \( y = s_{1} \oplus s_{2} \oplus \dots \oplus s_{t} \), where \( s_{i} \) represents the \( i \)-th intermediate result. LongDPO conducts learning at each step. Specifically, for the \((i+1)\)-th step, \( s_{w} \) is the chosen step, \( s_{l} \) is the rejected step, and \(s_{1 \sim i} = s_1\oplus \dots \oplus s_i\) has already been learned. LongDPO aims to maximize the probability of $s_{w}$ and minimize the probability of $s_{l}$.


\begin{multline} \label{eq:LongDPO}
    \mathcal{L}_{LongDPO} = -\mathbb{E}_{(q', s_{w}, s_{l}) \sim D} \Big[ \log \sigma \big( \\ \beta \log\frac{\pi_\theta(s_{w}|q')}{\pi_{ref}(s_{w}|q')}
    - \beta \log\frac{\pi_\theta(s_{l}|q')}{\pi_{ref}(s_{l}|q')} \big) \Big],
\end{multline}
where \( q' \) represents \( q \oplus s_{1 \sim i} \), which indicates the query concatenated with the corresponding steps learned up to the \( (i+1) \)-th step.

{
\begin{table*}[ht]
    \centering
    \small
    \begin{tabular}{l|cc|cc|cc|cc|cc cc}
    \toprule
      \multirow{2}{*}{Models} & \multicolumn{2}{c|}{\textbf{[0, 500)}} & \multicolumn{2}{c|}{\textbf{[500, 2k)}} & \multicolumn{2}{c|}{\textbf{[2k, 4k)}} & \multicolumn{2}{c|}{\textbf{[4k, 20k)}} & \multicolumn{2}{c}{\textbf{Average}} \\
     \cmidrule(lr){2-3} \cmidrule(lr){4-5} \cmidrule(lr){6-7} \cmidrule(lr){8-9} \cmidrule(lr){10-11} 
    & $S_l$ & $S_q$ & $S_l$ & $S_q$ & $S_l$ & $S_q$ & $S_l$ & $S_q$ & $S_l$ & $S_q$ \\
    \midrule
    \textbf{LongWriter-Llama} & 88.10 & 86.00 & 74.50 & 86.90 & 89.10 & 88.30 & 80.80 & 79.20 & 83.12 & 85.10 \\
    \qquad\emph{\textbf{w/ DPO}} & \textbf{90.93} & 85.78 & 76.67 & 85.46 & 90.01 & 90.53 & 81.07 & 80.90 & 85.55 & 85.66 \\
    \qquad\emph{\textbf{w/ LongDPO}} & 90.68 & \textbf{86.27} & \textbf{77.23} & \textbf{91.25} & \textbf{93.35} & \textbf{90.53} & \textbf{88.25} & \textbf{85.06} & \textbf{87.38} & \textbf{88.28}  \\
   \cmidrule(lr){1-1}  \cmidrule(lr){2-3} \cmidrule(lr){4-5} \cmidrule(lr){6-7} \cmidrule(lr){8-9} \cmidrule(lr){10-11} 
   \textbf{LongWriter-Qwen} & \textbf{90.80} & 87.99 & 84.37 & 89.37 & 84.21 & 84.84 & 58.69 & 78.13 & 79.51 & 85.08 \\
    \qquad\emph{\textbf{w/ DPO}} & 86.32 & 88.23 & \textbf{88.71} & 89.16 & \textbf{89.28} & 84.09 & 60.89 & 78.82 & 81.30 & 85.07 \\
    \qquad\emph{\textbf{w/ LongDPO}} & 88.93 & \textbf{91.91} & 85.47 & \textbf{91.25} & 88.63 & \textbf{85.60} & \textbf{71.14} & \textbf{85.41} & \textbf{83.54} & \textbf{88.54}  \\    
    \bottomrule
    \end{tabular}
    \caption{Evaluation results on LongBench-Write-en. LongWriter-Llama and LongWriter-Qwen represent LongWriter-llama-8B and LongWriter-Qwen2.5-7B. We have set a random seed to ensure reproducibility.}
    \label{main_result}
\end{table*}
}

\section{Experimental Results}
\subsection{Setting Up}
\label{exp:setup}
\paragraph{Setting on Collecting Stepwise Pair}
We conduct our experiments using LongWriter-llama3.1-8b~\footnote{\url{https://huggingface.co/THUDM/LongWriter-llama3.1-8b}} and LongWriter-Qwen2.5-7B-Instruct~\footnote{\url{https://www.modelscope.cn/models/swift/MS-LongWriter-Qwen2.5-7B-Instruct}}. To evaluate text rewards and generate critiques for Eq~\ref{critic-augmented}, we utilize Llama-3.1-70B-Instruct~\footnote{\url{https://huggingface.co/meta-llama/Llama-3.1-70B-Instruct}}.
For the MCTS tree configuration, we set the maximum depth to 4, with each node generating 4 child nodes during expansion. Each node can contain up to 2048 tokens, and we use a decoding temperature of 0.7, along with a fixed random seed for reproducibility.

\paragraph{Training Setting}
We randomly sample 2.5K instructions from WildChat~\citep{zhaowildchat} to collect stepwise preference pairs, which we then combine with UltraFeedback~\citep{DBLP:conf/icml/CuiY0YH0NXXL0024} for training. For data from UltraFeedback, we use vanilla DPO. The learning rate is set to 1e-6, with a cosine learning rate scheduler. 
The maximum sequence length is 32,768 through packing, with a random seed set to 42, and training for 250 steps. We use Xtuner~\footnote{\url{https://github.com/InternLM/xtuner}} for training.

\paragraph{Evaluation}
We evaluate long-form generation capabilities using the following benchmark:
\begin{itemize}
    \item \textbf{LongBench-Write} employs two metrics: the length score \( S_{l} \), which assesses how closely the model's generated length matches the required length, and the quality score \( S_{q} \), which evaluates the quality of the model's output using GPT-4o~\citep{bai2024longwriterunleashing10000word}. Our evaluation is performed using the English version.
    \item \textbf{LongGenBench}~\citep{wu2024spinning} evaluates whether models can maintain writing coherence and follow instructions which proposes three metrics to evaluate. 
    Completion Rate (CR) assesses the degree to which all designated subtasks are successfully completed. 
    STIC-1 evaluates the model’s adherence to
specific task instructions. 
STIC-2 provides more granular evaluations, measuring the overall completion of specific task instructions.
\end{itemize}
We use the official scripts for evaluation \footnote{\url{https://github.com/THUDM/LongWriter}} \footnote{\url{https://github.com/mozhu621/LongGenBench}}. 
Additionally, we assess the model's general abilities using the following: 
\begin{itemize}
  \item \textbf{TruthfulQA}~\citep{DBLP:conf/acl/LinHE22} to evaluate the helpfulness of the model's response.
  \item \textbf{MMLU}~\citep{hendryckstest2021} to evaluate the model's multitask processing. We use a 5-shot evaluation in our assessment following~\citep{grattafiori2024llama3herdmodels} setting.
  \item \textbf{GSM8K}~\citep{cobbe2021trainingverifierssolvemath} to evaluate the reasoning ability of LLM. We use an 8-shot evaluation following~\citep{grattafiori2024llama3herdmodels} setting.
\end{itemize}
We utilize UltraEval~\citep{he-etal-2024-ultraeval} and lm-evaluation-harness~\citep{eval-harness} for evaluation. 

\paragraph{Baselines}
The \textbf{LongWriter-(.) w/ DPO} baseline models are versions of \textbf{LongWriter-(.)} that have been trained using DPO. 
For each instruction from WildChat~\citep{zhaowildchat}, we generate four responses. The response with the highest reward is selected as the chosen candidate, while one of the remaining responses is randomly selected as the rejected candidate. Then combine UltraFeedback for training. 



{
\begin{table*}[ht]
\centering
\small
\begin{tabular}{lcccccccccc}
\toprule
\multirow{2}{*}{Models} & \multicolumn{3}{c}{\textbf{LongGenBench (16k)}} & \multicolumn{3}{c}{\textbf{LongGenBench (32k)}} & \multicolumn{2}{c}{\textbf{TruthfulQA}} & \textbf{MMLU} & \textbf{GSM8k} \\
\cmidrule(lr){2-4} \cmidrule(lr){5-7} \cmidrule(lr){7-7} \cmidrule(lr){8-9}  \cmidrule(lr){10-10}  \cmidrule(lr){11-11} 
 & CR & STC1 & STC2 & CR & STC1 & STC2 & ACC & ACC & ACC & ACC \\
\cmidrule(lr){1-1} \cmidrule(lr){2-4} \cmidrule(lr){5-7} \cmidrule(lr){7-7} \cmidrule(lr){8-9}   \cmidrule(lr){10-10}  \cmidrule(lr){11-11} 
\textbf{LongWriter-Llama} & 46.00 & 22.60 & 9.80  & 34.50 & 33.60 & 10.00 & 38.43 & 56.07 & 63.24 & 57.70 \\
\qquad\emph{\textbf{w/ DPO}}  & 64.99 & 25.99 & 16.29 & 65.24 & 32.47 & 20.39 & 38.17 & 55.68 & 63.30 & 59.20  \\
\qquad\emph{\textbf{w/ LongDPO}} & \textbf{69.38} & \textbf{27.59} & \textbf{18.45} & \textbf{68.35} & \textbf{33.69} & \textbf{22.15} & \textbf{40.76} & \textbf{58.78} & \textbf{63.67} & \textbf{61.30} \\
\cmidrule(lr){1-1}  \cmidrule(lr){2-4} \cmidrule(lr){5-7} \cmidrule(lr){7-7} \cmidrule(lr){8-9}   \cmidrule(lr){10-10}  \cmidrule(lr){11-11} 
\textbf{LongWriter-Qwen}  & \textbf{98.94} & 31.39 & 31.02 & 58.67 & \textbf{33.58} & 18.93 & \textbf{45.29} & 61.78 & 74.16 & 83.78 \\
\qquad\emph{\textbf{w/ DPO}}  & 95.95  & 31.18  & 29.83   & 82.23 & 29.02 & 22.33  & 39.29  & 57.67 & 63.67 & 83.85    \\
\qquad\emph{\textbf{w/ LongDPO}} & 98.51 & \textbf{33.07} & \textbf{32.52} & \textbf{84.95} & 29.86 & \textbf{24.32} & 44.92 & \textbf{62.75} & \textbf{74.25} & \textbf{84.08} \\
\bottomrule
\end{tabular}
\caption{Performance comparison across more long-form and general benchmarks. LongGenBench can be used to evaluate output lengths up to 32k. For TruthfulQA, we report partition ``MC1'' and ``MC2''. For each task, all three methods use the same decoding settings, and we have set a random seed to ensure reproducibility.}
\label{tab:long_general}
\end{table*}
}

{
\begin{table*}[ht]
    \centering
    \small
    \begin{tabular}{l|cc|cc|cc|cc|cc cc}
    \toprule
      \multirow{2}{*}{Methods} & \multicolumn{2}{c|}{\textbf{[0, 500)}} & \multicolumn{2}{c|}{\textbf{[500, 2k)}} & \multicolumn{2}{c|}{\textbf{[2k, 4k)}} & \multicolumn{2}{c|}{\textbf{[4k, 20k)}} & \multicolumn{2}{c}{\textbf{Average}} \\
     \cmidrule(lr){2-3} \cmidrule(lr){4-5} \cmidrule(lr){6-7} \cmidrule(lr){8-9} \cmidrule(lr){10-11} 
    & $S_l$ & $S_q$ & $S_l$ & $S_q$ & $S_l$ & $S_q$ & $S_l$ & $S_q$ & $S_l$ & $S_q$ \\
    \midrule
    \textbf{LongWriter-Llama} & 88.10 & 86.00 & 75.40 & 86.90 & 89.10 & 88.30 & 80.80 & 79.20 & 83.12 & 85.30 \\
    \qquad\emph{\textbf{w/o critique}} & 89.69 & 87.00 & 75.46 & 89.58 & 92.72 & 89.01 & 83.93 & 79.51 & 85.45 & 86.27 \\
    \qquad\emph{\textbf{w/ self-critique}} & \underline{92.51} & 88.15 & 74.40 & 89.81 & 90.15 & 88.48 & 83.62 & 81.38 & 85.17 & 86.96 \\
    \qquad\emph{\textbf{w/ LongDPO}} & 90.74 & \underline{89.14} & \underline{76.61} & \underline{90.70} & \underline{93.46} & \underline{91.10} & \underline{87.77} & \underline{81.94} & \textbf{87.14} & \textbf{88.22} \\
   \cmidrule(lr){1-1}  \cmidrule(lr){2-3} \cmidrule(lr){4-5} \cmidrule(lr){6-7} \cmidrule(lr){8-9} \cmidrule(lr){10-11}
    \textbf{LongWriter-Qwen} & \underline{90.80} & 87.99 & 84.37 & 89.37 & 84.21 & 84.84 & 58.69 & 78.13 & 79.51 & 85.08 \\
   \qquad\emph{\textbf{w/o critique}} & 89.59 & 86.99 & 85.35 & 89.01 & 88.14 & 84.31 & 63.98 & 80.20 & 81.77 & 85.12 \\
    \qquad\emph{\textbf{w/ self-critique}} & 90.67 & 90.68 & 83.60 & 
    \underline{93.26} & 87.46 & 86.61 & 65.20 & 78.24 & 81.73 & 87.20 \\
    \qquad\emph{\textbf{w/ LongDPO}} & 89.36 & \underline{91.18} & \underline{85.48} & 92.10 & \underline{89.60} & \underline{87.16} & 
    \underline{67.66} & \underline{83.17} & \textbf{83.03} & \textbf{88.40} \\ 
    \bottomrule
    \end{tabular}
    \caption{Ablation on refinement methods and ``w/o critique'' stands for without critiques meaning MCTS is applied alone. ``Self-critique'' refers to critiques generated by the model itself. To verify generalization, we set different values of \(\eta\) and report the average result.}
    \label{self-refine}
\end{table*}
}
\subsection{Main Results}
The main results are presented in Table~\ref{main_result}. Our method significantly outperforms baselines across both the Llama and Qwen series models. 
Consistent with the results of~\citet{bai2024longwriterunleashing10000word}, the use of DPO alone did not lead to a substantial performance improvement. 
This could be due to the challenge of maintaining response quality when directly sampling long responses generated by DPO~\citep{cheng2024sparselfplaytreesearchrefinement}. 
In contrast, our method demonstrates performance gains, likely because fine-grained supervision facilitates the acquisition of high-quality data.

To be specific, regarding the length score, LongWriter-Llama w/ LongDPO consistently shows improvements across various lengths, generating text that more accurately meets the length requirements. Notably, for outputs exceeding 4,000 words, performance improved by approximately 8\%. 
The quality score results are detailed in Table~\ref{tb:quality_detail}. 
When comparing LongWriter-Llama and LongWriter-Llama w/ DPO, 
the primary factors contributing to the improved scores of our generated texts are enhancements in ``Clarity," ``Breadth and Depth," and ``Reading Experience."

In addition to the 7B-sized model, we also conducted experiments on larger models and compared them with more advanced open-source models. Detailed results can be seen in Table~\ref{larger_models}.

\subsection{Generalization on more long-form and general benchmarks}
Table~\ref{tab:long_general} displays the results of various methods on LongGenBench. For both the Llama and Qwen series models, their performance on LongGenBench shows significant improvement. Notably, in terms of CR, this suggests that the model can better follow instructions after being trained with LongDPO. Additionally, using LongDPO results in better performance than DPO.

For other tasks, a similar trend can be observed: directly applying DPO fails to deliver significant performance improvements and, in some cases, even leads to notable declines. This is particularly evident in the MMLU task, where the performance of LongWriter-Qwen significantly deteriorates after applying DPO. In contrast, our method results in virtually no degradation of the model's other capabilities and even leads to slight improvements. This illustrates the generalizability of our approach to tasks beyond long-form generation.

\begin{table*}[ht]
\centering
\resizebox{\textwidth}{!}{
\begin{tabular}{lccccccccccc|c}
\toprule
           & \multicolumn{10}{c}{\textbf{LLM-AggreFact} (\emph{without} threshold tuning)}                                                     \\
\cmidrule(r){2-11} 
\multirow{2}{*}{\bf Model Name} &
  \multicolumn{2}{c}{\textbf{\textsc{AggreFact}}} &
  \multicolumn{2}{c}{\textbf{\textsc{TofuEval}}} &
  \multirow{2}{*}{\textbf{\textsc{Wice}}} &
  \multirow{2}{*}{\textbf{\textsc{Reveal}}} &
  \multirow{2}{*}{\begin{tabular}[c]{@{}c@{}}\textbf{\textsc{Claim}}\\ \textbf{\textsc{Verify}}\end{tabular}} &
  \multirow{2}{*}{\begin{tabular}[c]{@{}c@{}}\textbf{\textsc{Fact}}\\ \textbf{\textsc{Check}}\end{tabular}} &
  \multirow{2}{*}{\begin{tabular}[c]{@{}c@{}}\textbf{\textsc{Expert}}\\ \textbf{\textsc{QA}}\end{tabular}} &
  \multirow{2}{*}{\textbf{\textsc{Lfqa}}} &
  \multirow{2}{*}{\textbf{\textsc{RT}}} &
  \multirow{2}{*}{\textbf{Avg}} \\
\cmidrule(r){2-3} \cmidrule(r){4-5} 
& \textbf{CNN}  & \textbf{XSum} & \textbf{MediaS} & \textbf{MeetB} &      &      &      &      &      &    &  \\
\midrule
\quad \textit{LongWriter-Qwen} & 52.71 & 71.55 & \underline{73.33} & 75.83 & 74.40 & 87.73 & 70.18 & 74.61 & 60.56 & 84.61 & 76.65 & 72.92 \\
\quad \textit{w/o Memory} & 52.03 & 69.31 & 72.16 & 75.38 & \underline{76.07} & 87.58 & 68.46 & \underline{74.94} & 60.27 & 83.36 & 75.70 & 72.30 \\
\quad \textit{w/ Memory} & \underline{54.36} & \underline{73.20} & 73.28 & \underline{76.25} & 74.92 & \underline{88.31} & \underline{70.87} & 73.79 & \underline{61.23} & \underline{86.76} & \underline{77.39} & \textbf{73.67} \\
\bottomrule
\end{tabular}
} 
\caption{Performance (BAcc) of evaluator models on the test split of LLM-AggreFact. ``RT'' represents RAGTruth.
}
\label{tab:res-llm-aggrefact}
\end{table*}

\subsection{Comparision with Different Critic Methods}
Self-critique is widely used~\citep{ankner2024critiqueoutloudrewardmodels, ye2024improvingrewardmodelssynthetic} to leverage models' internal knowledge to provide feedback to provide a better solution. 
However, recent studies have emphasized that relying solely on a model's internal knowledge can result in unstable performance gains~\citep{qi2024mutualreasoningmakessmaller, zhang2024understandingdarkllmsintrinsic}. 
To further verify whether self-generated critiques can effectively collect better preference pairs, we compare self-generated critiques with external critiques in Table~\ref{self-refine}. We have ensured that the only difference lies in the critic model used between self-critique and LongDPO.

To enable a more thorough comparison, we set multiple values for \(\eta\) in Equation~\ref{thresh_hold}. Specifically, we set \(\eta\) to \{2.0, 2.5, 3.0\} and report the average performance in Table~\ref{self-refine}. We detailed the results in Table~\ref{sigma} and~\ref{sigma_qwen}. Self-critique exhibits performance fluctuations which may be because the generator's internal knowledge is insufficient, making it difficult to distinguish high-quality steps.

\subsection{Effects of the Memory Pool}
We assess the effectiveness of the memory pool using the LLM-AggreFact~\citep{tang-etal-2024-minicheck}, which includes a variety of fact-checking tasks. The results are presented in Table~\ref{tab:res-llm-aggrefact}. Without using memory to collect data and training directly, the fact-checking scores decreased. However, after incorporating memory, the model's fact-checking ability improved.

{
\begin{table}[ht]
\centering
\small
\resizebox{0.9\columnwidth}{!}{
\begin{tabular}{lccc}
\toprule
\multirow{2}{*}{Models} & \multicolumn{3}{c}{\textbf{LongGenBench}}  \\
\cmidrule(lr){2-4} 
 & CR & STC1 & STC2 \\
\cmidrule(lr){1-4} 
\rowcolor{gray!20}  \multicolumn{4}{l}{\textit{LongWriter-Llama}} \\
\qquad\emph{\textbf{w/o Stepwise}} & 67.89 & 25.36 & 17.29 \\
\qquad\emph{\textbf{w/ Stepwise}} & \textbf{69.38} & \textbf{27.59} & \textbf{18.45} \\
\cmidrule(lr){1-4} 
\rowcolor{gray!20}  \multicolumn{4}{l}{\textit{LongWriter-Qwen}} \\
\qquad\emph{\textbf{w/o Stepwise}} & 97.42  & 31.95  & 31.44     \\
\qquad\emph{\textbf{w/ Stepwise}} & \textbf{98.51} & \textbf{33.07} & \textbf{32.52}  \\
\bottomrule
\end{tabular}
}
\caption{Performance comparison in LongGenBench.}
\label{tab:loss_ablation}
\end{table}
}

\subsection{Effects of Stepwise Learning}
We evaluate the impact of stepwise learning on long-form generation using LongGenbench. The results are shown in Table~\ref{tab:loss_ablation}. We use the same training data. The difference between the methods is that ``w/o Stepwise'' refers to training with vanilla DPO, while ``w/ Stepwise'' refers to training with the LongDPO objective. Stepwise learning is beneficial for learning long-form generation. The detailed results shown in Table~\ref{tab:loss_ablation_full}.

\section{Analysis}
\subsection{Reliability of Evaluation}

\begin{table}[ht]
\centering
\resizebox{0.45\textwidth}{!}{%
\begin{tabular}{lccc}
\toprule
\textbf{Rate} & \textbf{Diversity} & \textbf{Consistency} & \textbf{Informative} \\
\midrule
Win & 65.0 & 61.7 & 61.7 \\
Tie & 8.30 & 16.7 & 6.70 \\
Lose & 26.7 & 21.6 & 31.6 \\
\bottomrule
\end{tabular}
}
\caption{Human evaluation with win rates under three criteria: Diversity, Consistency, and Informativeness}
\label{tab:human_result}
\end{table}

\textbf{Reliability on Quality Score} 
We evaluate the consistency of GPT-4o in LongBench-Write based on three evaluation runs and report the variance following~\citep{longwriter_openreview}. Table~\ref{tab:models_sq} presents the results of the average quality score, which may indicate that GPT-4o demonstrates good consistency.

\begin{table}[ht]
\centering
\resizebox{0.45\textwidth}{!}{%
\begin{tabular}{lccc}
\toprule
 \textbf{Judge} & \textbf{Judge-1} & \textbf{Judge-2} & \textbf{Judge-3} \\
\midrule
Judge-1 & - & 61.7 & 63.4 \\
Judge-2 & 61.7 & - & 61.7 \\
Judge-3 & 65.0 & 58.4 & - \\
\bottomrule
\end{tabular}
}
\caption{Human agreement between different annotators. Judge-1, Judge-2, and Judge-3 are three human judges.}
\label{tab:human_agreement}
\end{table}

\textbf{Human Evaluation} 
In addition to utilizing GPT-4o, we conduct a human evaluation to assess the generated text in terms of diversity, consistency, and informative detailed guidelines can be seen in~\ref{human_annotation}.
We compare the responses generated by LongWriter-Llama and LongWriter-Qwen with those produced by the same models trained using LongDPO.
Three independent annotators, who are undergraduate and graduate students, are tasked with comparing the response pairs and evaluating them as win, tie, or lose. The student participants all possess a bachelor's or master's degree and are from top universities and have two years of experience in NLP.
The results, present in Table~\ref{tab:human_result}, indicate that our responses are rated as superior by the human judges. 
Additionally, Table~\ref{tab:human_agreement} shows the agreement among the three judges, demonstrating a high level of consistency in their evaluations.

\subsection{Case Study} 
Figure~\ref{case_study} presents a case sampled from LongGenBench. The instruction primarily requires visiting the farmers' market starting from week 10 and then every 5 weeks thereafter. LongWriter-Llama fulfills the requirement in week 10 but fails in week 15. However, after applying LongDPO, it is able to consistently meet the demands.

We analyze the attention distribution across models and observe that, in week 15, LongWriter-Llama fails to attend to ``farmers market.'' However, after applying LongDPO, it successfully does so. We find that a small number of attention heads have attended to ``farmers market,'' with over 1\% of attention heads scoring above 0.5. However, the LongWriter model does not exhibit a similar pattern. This behavior may be linked to retrieval heads~\citep{wu2024retrievalheadmechanisticallyexplains}. We also provide examples in Figure~\ref{fact1} and~\ref{fact2} to show factual correctness after applying LongDPO. 

\section{Disscussion}
LongDPO focuses on long-form tasks (e.g., Creative Writing), which, unlike tasks such as math and coding, do not have a ground truth. It is more challenging to assess the reward precisely. Different from existing literature in reinforcement learning, which can rely on rule-based rewards or process reward models, we take into full consideration the characteristics of natural language and have carefully designed seven principles for evaluating the reward.

\section{Conclusion}
In this paper, we propose LongDPO which incorporate process supervision with MCTS to collect better preference pairs with a memory pool to maintain factual consistency and leverages external critiques to refine low-quality candidates in long-form generation. LongDPO enhances performance in long-form generation tasks (e.g.~LongBench-Write) while maintaining near-lossless performance on several general tasks.

\section*{Limitations}
We have validated the effectiveness of LongDPO in generating text of 32K length. However, due to the limitations of current benchmarks, it is challenging to evaluate longer generation lengths. In the future, we plan to test the performance of LongDPO further on longer benchmarks.

\section*{Acknowledgements}
This work was supported by the National Science and Technology Major Project (No. 2022ZD0117800).

\bibliography{acl_latex}


\appendix

\onecolumn
\section{Templates and Guidelines}
\subsection{Reward Evaluation Templates}
\begin{tcolorbox}
[colframe=blue!75!black,colback=blue!5!white,title=Reward Evaluation Template]
You are an expert at evaluating the quality of text.

As an impartial evaluator, please assess the assistant’s response to a user’s requirements. Now, you will receive specific principles that provide the criteria for evaluating the response. Principles begin,

\textbf{Principle1}: The response is accurate and free of factual errors.

\textbf{Principle2}: The response meets the user’s purpose and needs.

\textbf{Principle3}: The response is non-toxic and safe.

\textbf{Principle4}: The response meets the user’s formatting requirements and maintains logical consistency.

\textbf{Principle5}: The response contains diverse and comprehensive information with minimal repetition.

\textbf{Principle6}: The response provides an excellent reading experience.

\textbf{Principle7}: The response is insightful and provides the user with additional avenues for thought. Principles end.

In the next, you will receive detailed guidelines to help you rate the response according to each principle. Now, guidelines begin

\textbf{5}: A perfect response with no improvement needed. The content is comprehensive, accurate, clear, and well-structured. The response fully addresses all aspects of the question or need without any omissions or errors.

\textbf{4}: A very good response with minor issues. It is almost perfect but may have slight areas that could be improved, such as minor details that are unclear or a small omission. Overall, it still meets the need effectively.

\textbf{3}: An acceptable response that generally meets the question or need but has noticeable shortcomings. The content might be incomplete or unclear, or there may be minor grammar or logical errors. It needs improvement but is still functional.

\textbf{2}: A response with significant issues that requires substantial improvement. The content is incomplete, unclear, or contains major errors, omissions, or misunderstandings. It does not fully satisfy the request.

\textbf{1}: A completely inadequate response that fails to meet the question or need. It contains serious errors or misunderstandings and cannot provide useful help. 

Guidelines end. 

Now, you will receive the user request and the assistant's response to evaluate. 

\textbf{\textless User Request\textgreater}

\$INST\$

\textbf{\textless/User Request\textgreater}

\textbf{\textless Response\textgreater}

\$RESPONSE\$

\textbf{\textless/Response\textgreater}

Your task is to evaluate the quality of the response and assign a rating with distinguishable differentiation for each principle. When rating, please carefully read the guidelines and ensure your ratings fully adhere to them. You must first provide a brief analysis of its quality, then determine the weights for each \textbf{Principle}, for example \{"Principle1": [0.2,0.2,0.2,0.2,0.2]\} represents the final score is 0.2 * 1 + 0.2 * 2 + 0.2 * 3 + 0.2 * 4 + 0.2 * 5 = 3. The output must strictly follow the JSON format: {"Analysis":..., "Principle1":[..,..,..,..,..], "Principle2":[..,..,..,..,..], "Principle3":[..,..,..,..,..], "Principle4":[..,..,..,..,..], "Principle5":[..,..,..,..,..], "Principle6":[..,..,..,..,..], "Principle7":[..,..,..,..,..]}. You do not need to consider whether the response meets the user's length requirements in your evaluation. Ensure that only one integer or float is output for each principle.
\label{text_reward}
\end{tcolorbox}

\subsection{Templates for Generate Critiques}
\begin{tcolorbox}[colframe=blue!75!black, colback=blue!5!white, coltitle=white, title=Templates for Generate Critiques]
You are an expert at evaluating the quality of text. In the following, you will revice a user request, one principle and two candidates:

\textbf{\textless User Request\textgreater}

\$INST\$

\textbf{\textless/User Request\textgreater}

\textbf{\textless Principle\textgreater}

\$PRINCIPLE\$

\textbf{\textless/Principle\textgreater}

\textbf{\textless Candidate1\textgreater}

\$CANDIDATE1\$

\textbf{\textless/Candidate1\textgreater}

\textbf{\textless Candidate2\textgreater}

\$CANDIDATE2\$

\textbf{\textless/Candidate2\textgreater}

Now, your task is 
1. Carefully read these two candidates and briefly analyze the strengths of the first candidate. 
2. Provide a "Justification" explaining why it scores higher. 
3. Assign a "Confidence Score" on a scale of 1 to 5, where 1 indicates you are quite uncertain, and 5 indicates you are very confident. 
4. Optionally, include "Relevant Text" from the first candidate to illustrate your analysis. 
5. Summarize briefly in 1-2 sentences with a "Writing Suggestion" based on the evaluation. The output must strictly follow the JSON format: \texttt{\{"Analysis":..., "Justification":..., "Writing Suggestion":..., "Confidence Score":...,"Relevant Text":...\}}. Ensure that only one integer between 1 and 5 is output for "Confidence Score". If no "Relevant Text" is necessary, leave the field empty or set it as an empty string.
\label{refine_template}
\end{tcolorbox}

\subsection{Templates for Check Consistency}
\begin{tcolorbox}[colframe=blue!75!black, colback=blue!5!white, coltitle=white, title=Template for Finding Fact]
You're an expert in natural language processing and information retrieval. You will receive a response. Your task is to extract factual statements from the response provided. 

Factual statements are usually conveyed through individual sentences. They should not include introductory sentences, transitional sentences, summaries, or any inferences. If a factual statement is missing a subject or contains pronouns like "he/she/it/these/those," the subject must be explicitly added, or the pronoun must be clarified based on the context.

Now, please process the following AI assistant’s response:

\textbf{\textless Response\textgreater}

\$RESPONSE\$

\textbf{\textless/Response\textgreater}

Please carefully read and analyze the given content. Then, breaking the factual content. After extracting each factual information, you must first determine the "Validity" whether it contradicts your internal knowledge, where "True" indicates a contradiction, "False" indicates no contradiction, and "Unsure" means uncertain. Provide the relevant "Evidence" accordingly. Then, output the result in the following format: \texttt{\{"Analysis":..., "Fact1":\{"Content":...,"Validity":...,"Evidence":...\}, "Fact2":\{"Content":...,"Validity":...,"Evidence":...\},...\}}. Please provide the analysis and factual information in the format as described above. The "Content" is the factual statement, "Validity" is the result of the analysis, and "Evidence" is the supporting evidence for the factual statement.
\label{find_fact}
\end{tcolorbox}

\begin{tcolorbox}[colframe=blue!75!black, colback=blue!5!white, coltitle=white, title=Template for Judge Inconsistency]
You are an expert at evaluating text. You will receive factual statements along with a related response. Your task is to carefully evaluate whether the response contradicts the factual statement. Please use the following principles to generate your assessment:

\textbf{Contradict}: You can find strong evidence indicating factual inaccuracies in the response that are inconsistent with the given factual statement.

\textbf{Not Contradict}: You are unable to find evidence indicating factual inaccuracies in the provided response that contradicts the given factual statement.
Ensure that you do not use any information or knowledge beyond the response provided, and only check whether the statement is supported by the response.

Now, please refer to the principles to give your judgement:

\textbf{\textless Statement\textgreater}

\$STATEMENT\$

\textbf{\textless/Statement\textgreater}

\textbf{\textless Response\textgreater}

\$RESPONSE\$

\textbf{\textless/Response\textgreater}

You must provide an analysis first, followed by the judgement. The output must strictly follow the JSON format: \texttt{\{"Analysis":..., "Judgement":...,"Evidence":...\}.}

\label{judge_fact}
\end{tcolorbox}

\subsection{Guidelines for Human Annotation}

\begin{tcolorbox}[colframe=blue!75!black, colback=blue!5!white, coltitle=white, title=Guidelines for Human Annotation]
    \textbf{1. Diversity:} Which text is more diverse in content? This can be evaluated holistically, considering factors such as the lexical variety, the richness of semantics, the complexity of writing style, and the diversity in article structure. \vspace{1em}

    \textbf{2. Consistency:} Which text demonstrates a higher degree of consistency? This can be assessed holistically, considering factors such as thematic coherence, ensuring the central theme remains clear; logical coherence, reflected in the natural flow of ideas; and factual consistency, verified through accurate and reliable information. \vspace{1em}

    \textbf{3. Informative:} Which text is more informative in content? This can be evaluated holistically, considering factors such as the accuracy of the information presented, the comprehensiveness in covering all relevant aspects, the clarity of explanations, and the ease of readability and understanding.
    \label{human_annotation}
\end{tcolorbox}

\begin{figure*}[ht]
  \centering
  \includegraphics[width=0.99\linewidth]{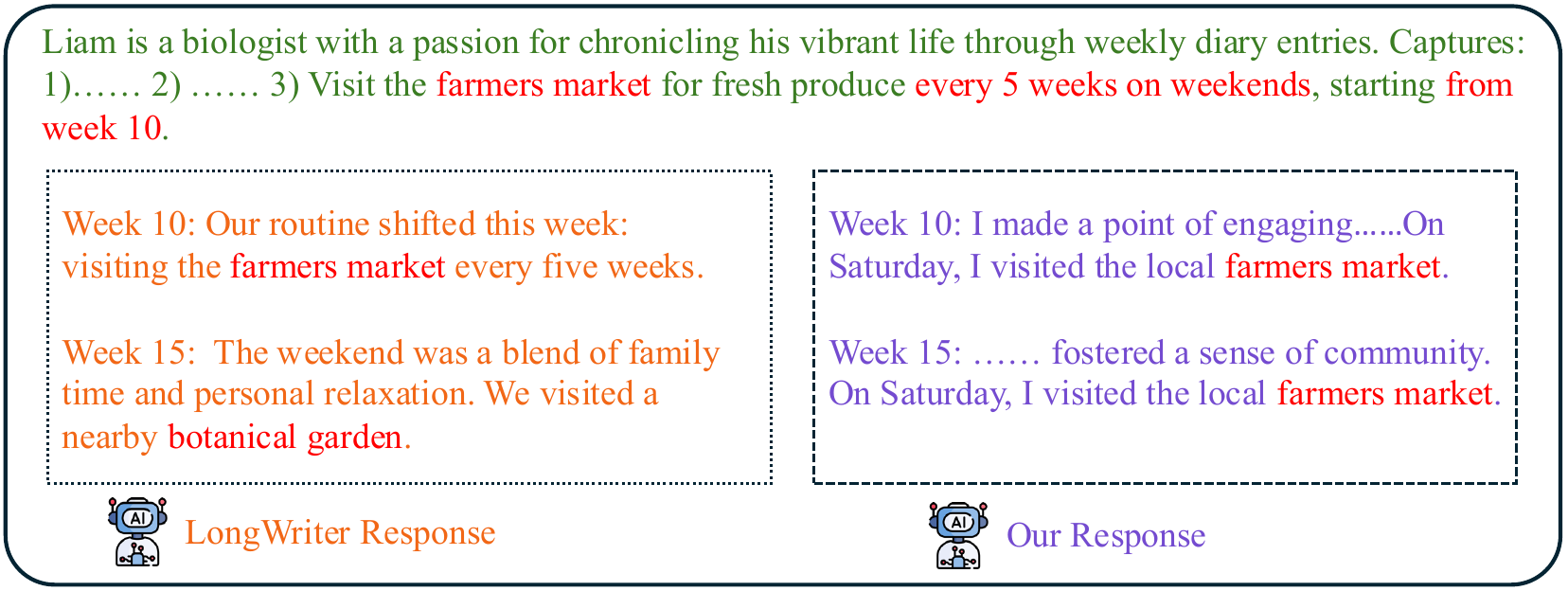}
  \caption{
  A case is randomly sampled from LongGenBench. The instruction primarily requires visiting the farmers' market starting from week 10 and then every 5 weeks thereafter. On the left, LongWriter-Llama fulfills the requirement in week 10 but fails in week 15. On the right, after applying LongDPO, LongWriter-Llama is able to consistently meet the demands.
  }
  \label{case_study}
\end{figure*}

\section{More Evaluation Results}

\begin{figure}[ht]
  \centering
  \includegraphics[width=0.5\linewidth]{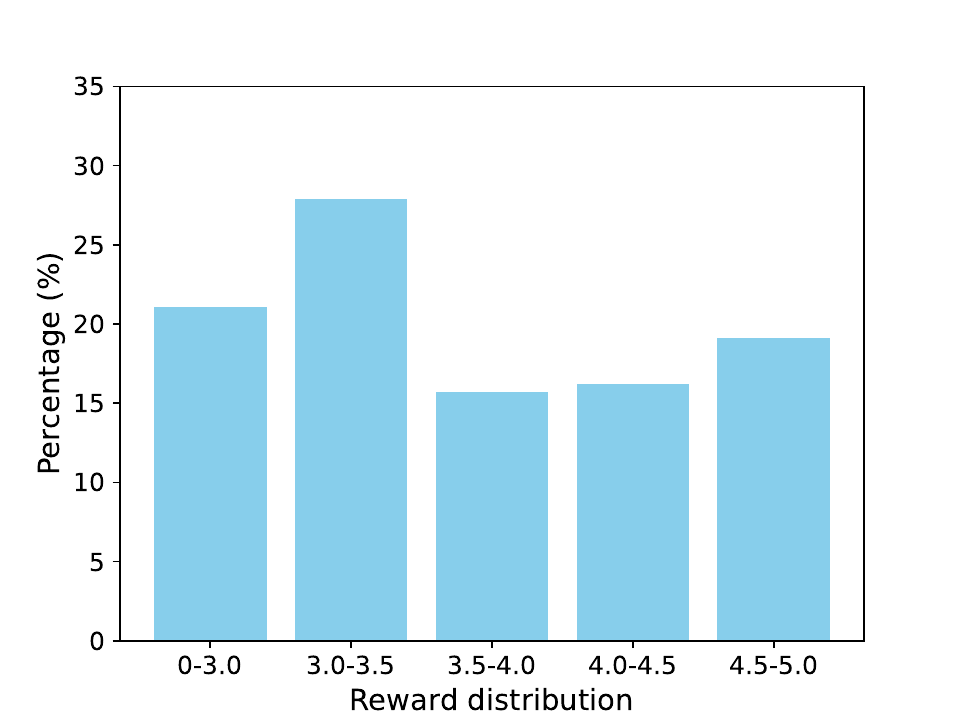}
  \caption{
  Reward analysis of the selected candidates, we focus solely on the chosen candidate in each preference pair. On the x-axis, '0-3.0' represents the proportion of candidates with an average reward  $ < 3.0$, while '3.0-3.5' represents the proportion of candidates with an average reward $\geq 3.0$ but $< 3.5$. Detailed reward distribution can be found in Appendix~\ref{reward_distribution_full}. 
  }
  \label{reward_distribution}
\end{figure}

\begin{table}[htbp]
    \centering
    \resizebox{0.9\linewidth}{!}{
    \begin{tabular}{l|ccccccc}
    \toprule
    & $S_q$ & Relevance & Accuracy & Coherence & Clarity & Breadth and Depth & Reading Experience \\
    \midrule
    LongWriter-Llama & 79.20 & 90.90 & 87.50 & 84.48 & 81.89 & 59.48 & 71.55 \\
    \quad\emph{+DPO} & 80.90 & 93.75 & 83.33 & 77.08 & 77.08 & 83.33 & 70.83 \\
    \quad\emph{+LongDPO} & 85.06 & 93.75 & 85.42 & 85.42 & 81.25 & 87.50 & 77.08 \\
    \midrule
    LongWriter-Qwen & 78.13 & 83.33 & 81.25 & 83.33 & 77.08 & 68.75 & 75.00 \\
    \quad\emph{+DPO} & 78.81 & 85.41 & 81.25 & 83.33 & 81.25 & 85.41 & 70.83 \\
    \quad\emph{+LongDPO} & 85.41 & 91.67 & 91.67 & 83.33 & 83.33 & 83.33 & 79.16 \\
    \bottomrule
    \end{tabular}
    }
    \caption{Detailed quality score for length exceeding 4000 in LongBench-Write-en.}
    \label{tb:quality_detail}
\end{table}

{
\setlength{\tabcolsep}{0.45em}
\begin{table*}[ht]
    \centering
    \small
    \begin{tabular}{l|cc|cc|cc|cc|cc}
    \toprule
      \multirow{2}{*}{LongWriter-Llama} & \multicolumn{2}{c|}{\textbf{[0, 500)}} & \multicolumn{2}{c|}{\textbf{[500, 2k)}} & \multicolumn{2}{c|}{\textbf{[2k, 4k)}} & \multicolumn{2}{c|
      }{\textbf{[4k, 20k)}} & \multicolumn{2}{c
      }{\textbf{Average}} \\
     \cmidrule(lr){2-3} \cmidrule(lr){4-5} \cmidrule(lr){6-7} \cmidrule(lr){8-9} \cmidrule(lr){10-11} 
    & $S_l$ & $S_q$ & $S_l$ & $S_q$ & $S_l$ & $S_q$ & $S_l$ & $S_q$ & $S_l$ & $S_q$  \\
    \midrule
    Self-critique \emph{+$ \eta \le 2.0$}  & 94.07 & 88.97 & 72.39 & 87.99 & 86.86 & 89.39 & 82.72 & 80.55 & 84.01 & 86.72 \\
     \qquad \emph{+$\eta \le 2.5$} & 93.08 & 88.48 & 76.43 & 91.04 & 91.66 & 88.54 & 84.63 & 82.35 & \textbf{86.45} & \textbf{87.60} \\
     \qquad \emph{+$\eta \le 3.0$} & 90.38 & 87.01 & 74.37 & 90.41 & 91.94 & 87.50 & 83.50 & 81.25 & 85.04 & 86.54 \\
   \cmidrule(lr){1-1}  \cmidrule(lr){2-3} \cmidrule(lr){4-5} \cmidrule(lr){6-7} \cmidrule(lr){8-9} \cmidrule(lr){10-11}
    LongDPO \emph{+$\eta \le 2.0$} & 92.01 & 92.91 & 72.55 & 91.45 & 93.35 & 93.75 & 88.86 & 80.20 & 86.69 & \textbf{89.57} \\
    \qquad \emph{+$\eta \le 2.5$} & 90.68 & 86.27 & 77.23 & 91.25 & 93.35 & 90.53 & 88.25 & 85.06 & \textbf{87.38} & 88.19 \\
    \qquad \emph{+$\eta \le 3.0$}  & 89.51 & 88.23 & 80.04 & 89.39 & 93.68 & 89.01 & 86.19 & 80.55 & 86.47 & 86.80 \\
    \bottomrule
    \end{tabular}
    \caption{Results on changing \(\eta\) using llama-based backbones}
    \label{sigma}
\end{table*}
}

{
\setlength{\tabcolsep}{0.45em}
\begin{table*}[ht]
    \centering
    \small
    \begin{tabular}{l|cc|cc|cc|cc|cc}
    \toprule
      \multirow{2}{*}{LongWriter-Qwen} & \multicolumn{2}{c|}{\textbf{[0, 500)}} & \multicolumn{2}{c|}{\textbf{[500, 2k)}} & \multicolumn{2}{c|}{\textbf{[2k, 4k)}} & \multicolumn{2}{c|}{\textbf{[4k, 20k)}} & \multicolumn{2}{c}{\textbf{Average}} \\
     \cmidrule(lr){2-3} \cmidrule(lr){4-5} \cmidrule(lr){6-7} \cmidrule(lr){8-9} \cmidrule(lr){10-11} 
    & $S_l$ & $S_q$ & $S_l$ & $S_q$ & $S_l$ & $S_q$ & $S_l$ & $S_q$ & $S_l$ & $S_q$ \\
    \midrule
    Self-critique \emph{+$\eta \le 2.0$} & 88.71 & 88.23 & 84.45 & 93.54 & 86.37 & 84.46 & 64.88 & 78.47 & 81.10 & 86.17 \\
   \qquad \emph{+$\eta \le 2.5$} & 91.96 & 91.66 & 83.16 & 92.91 & 88.94 & 86.36 & 67.69 & 79.16 & \textbf{82.93} & 87.52 \\
   \qquad \emph{+$\eta \le 3.0$} & 91.33 & 92.15 & 83.20 & 93.33 & 87.06 & 89.01 & 63.04 & 77.08 & 81.16 & \textbf{87.89} \\
   \cmidrule(lr){1-1}  \cmidrule(lr){2-3} \cmidrule(lr){4-5} \cmidrule(lr){6-7} \cmidrule(lr){8-9} \cmidrule(lr){10-11}
    LongDPO \emph{+$\eta \le 2.0$} & 87.84 & 91.45 & 86.21 & 92.15 & 91.35 & 86.86 & 66.85 & 82.59 & 83.06 & 88.26 \\
    \qquad \emph{+$\eta \le 2.5$} & 88.93 & 91.91 & 85.47 & 91.25 & 88.63 & 85.60 & 71.14 & 85.41 & \textbf{83.54} & \textbf{88.54} \\
    \qquad \emph{+$\eta \le 3.0$} & 91.32 & 90.19 & 84.75 & 92.91 & 88.82 & 89.01 & 64.99 & 81.51 & 82.47 & 88.51 \\
    \bottomrule
    \end{tabular}
    \caption{Results on changing $\eta$ using Qwen-based backbones}
    \label{sigma_qwen}
\end{table*}
}

{
\begin{table}[ht]
\centering
\small
\resizebox{0.7\columnwidth}{!}{
\begin{tabular}{lcccccc}
\toprule
\multirow{2}{*}{Models} & \multicolumn{3}{c}{\textbf{LongGenBench (16K)}} & \multicolumn{3}{c}{\textbf{LongGenBench (32K)}} \\
\cmidrule(lr){2-4} \cmidrule(lr){5-7} 
 & CR & STC1 & STC2  & CR & STC1 & STC2  \\
\cmidrule(lr){1-1} \cmidrule(lr){2-4} \cmidrule(lr){5-7} 
\rowcolor{gray!20}  \multicolumn{7}{l}{\textit{LongWriter-Llama}} \\
\qquad\emph{\textbf{w/o Stepwise}} & 67.89 & 25.36 & 17.29 & 67.79 & 31.85 & 21.67 \\
\qquad\emph{\textbf{w/ Stepwise}} & \textbf{69.38} & \textbf{27.59} & \textbf{18.45} & \textbf{68.35} & \textbf{33.69} & \textbf{22.15} \\
\cmidrule(lr){1-7} 
\rowcolor{gray!20}  \multicolumn{7}{l}{\textit{LongWriter-Qwen}} \\
\qquad\emph{\textbf{w/o Stepwise}} & 97.42  & 31.95  & 31.44 & 83.78 & 28.82 & 23.24   \\
\qquad\emph{\textbf{w/ Stepwise}} & \textbf{98.51} & \textbf{33.07} & \textbf{32.52} & \textbf{84.95} & \textbf{29.86} & \textbf{24.32}  \\
\bottomrule
\end{tabular}
}
\caption{Performance comparison in LongGenBench.}
\label{tab:loss_ablation_full}
\end{table}
}

\begin{table}[ht]
\centering
\begin{tabular}{lcc}
    \toprule
    \textbf{Evaluated Models} & \boldmath$S_q$ \\
    \midrule
    Claude 3.5 Sonnet & $87.7 \pm 0.5$ \\
    GPT-4 Turbo       & $86.6 \pm 0.4$ \\
    GPT-4o mini       & $90.3 \pm 0.3$ \\
    GPT-4o            & $91.8 \pm 0.5$ \\
    GLM-4-9B-chat     & $85.5 \pm 0.4$ \\
    Llama-3.1-8B-Instruct & $70.6 \pm 0.3$ \\
    Llama-3.1-70B-Instruct & $80.3 \pm 0.3$ \\
    Mistral-Large-Instruct & $88.3 \pm 0.4$ \\
    Suri-I-ORPO       & $53.5 \pm 0.5$ \\
    LongWriter-Llama     & $82.2 \pm 0.4$ \\
    LongWriter-Llama + LongDPO   & $88.2 \pm 0.5$ \\
    LongWriter-Qwen + LongDPO   & $88.6 \pm 0.5$ \\
    \bottomrule
\end{tabular}
\caption{Evaluated Models and the average $S_q$ Scores. We evaluate LongWriter-Llama + LongDPO and LongWriter-Qwen + LongDPO, while \citet{longwriter_openreview} report the remaining results.}    
\label{tab:models_sq}
\end{table}

{
\begin{table*}[ht]
    \centering
    \small
    \begin{tabular}{c|cc|cc|cc|cc|cc cc}
    \toprule
      \multirow{2}{*}{Models} & \multicolumn{2}{c|}{\textbf{[0, 500)}} & \multicolumn{2}{c|}{\textbf{[500, 2k)}} & \multicolumn{2}{c|}{\textbf{[2k, 4k)}} & \multicolumn{2}{c|}{\textbf{[4k, 20k)}} & \multicolumn{2}{c}{\textbf{Average}} \\
     \cmidrule(lr){2-3} \cmidrule(lr){4-5} \cmidrule(lr){6-7} \cmidrule(lr){8-9} \cmidrule(lr){10-11} 
    & $S_l$ & $S_q$ & $S_l$ & $S_q$ & $S_l$ & $S_q$ & $S_l$ & $S_q$ & $S_l$ & $S_q$ \\
    \midrule
    \textbf{Llama3.1-8B-instruct} & 89.70 & 84.60 & 78.20 & 80.60 & 29.20 & 76.10 & 0 & 57.60 & 56.80 & 76.30 \\
    \textbf{Llama3.1-70B-instruct} & 90.80 & 84.80 & 88.60 & 84.40 & 14.90 & 84.50 & 0 & 78.00 & 59.00 & 80.30 \\
    \textbf{GPT-4o} & 92.10 & 93.10 & 92.20 & 93.50 & 53.00 & 92.80 & 6.20 & 81.20 & 67.80 & 90.90  \\
   \textbf{QWQ} & 89.10 & 94.58 & 90.59 & 94.31 & 33.13 & 93.75 & 0.26 & 89.39 & 53.27 & \textbf{93.01} \\
    \cmidrule(lr){1-1}  \cmidrule(lr){2-3} \cmidrule(lr){4-5} \cmidrule(lr){6-7} \cmidrule(lr){8-9} \cmidrule(lr){10-11} 
    \textbf{Qwen-2.5-14B} & 88.39 & 89.77 & 81.83 & 91.91 & 71.38 & 87.50 & 19.57 & 82.35 & 65.29 & 87.88  \\
    \qquad\emph{\textbf{w/ DPO}} & 91.72 & 90.53 & 81.79 & 92.72 & 68.94 & 86.50 & 18.33 & 80.20 & 65.19 & 87.48  \\
    \qquad\emph{\textbf{w/ LongDPO}} & 91.75 & 91.25 & 85.69 & 90.53 & 78.79 & 89.01 & 21.50 & 86.04 & \textbf{69.43} & 89.21  \\
    \bottomrule
    \end{tabular}
    \caption{More evaluation results of larger models on LongBench-Write-en. }
    \label{larger_models}
\end{table*}
}

\begin{figure}[ht]
  \centering
  \includegraphics[width=0.50\linewidth]{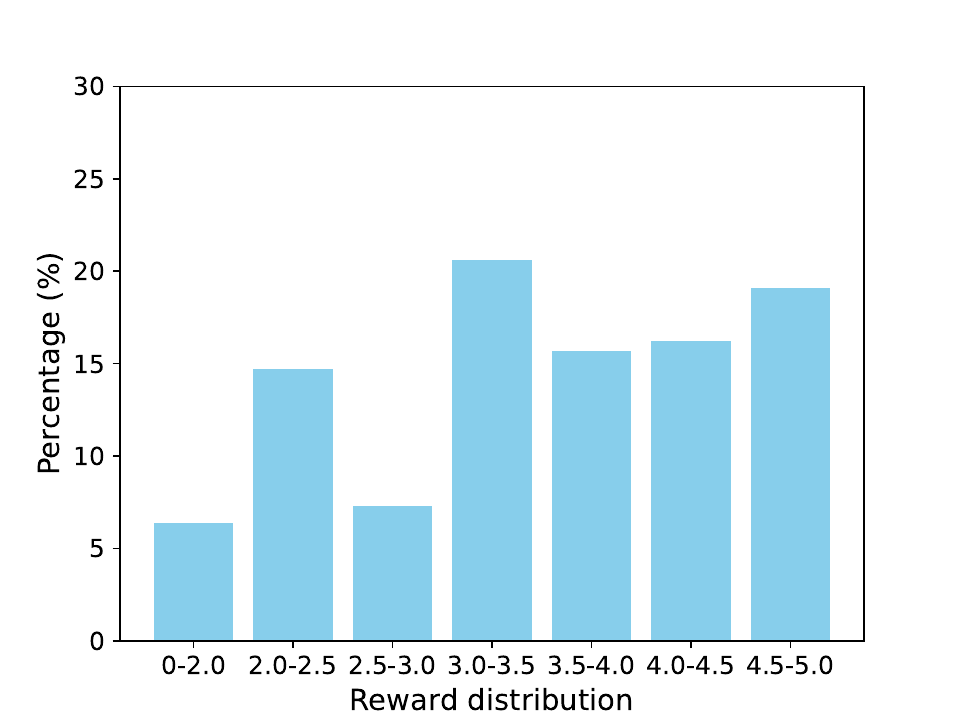}
  \caption{
  Detailed reward analysis of the chosen candidates.
  }
  \label{reward_distribution_full}
\end{figure}

\begin{figure}[ht]
  \centering
  \includegraphics[width=0.50\linewidth]{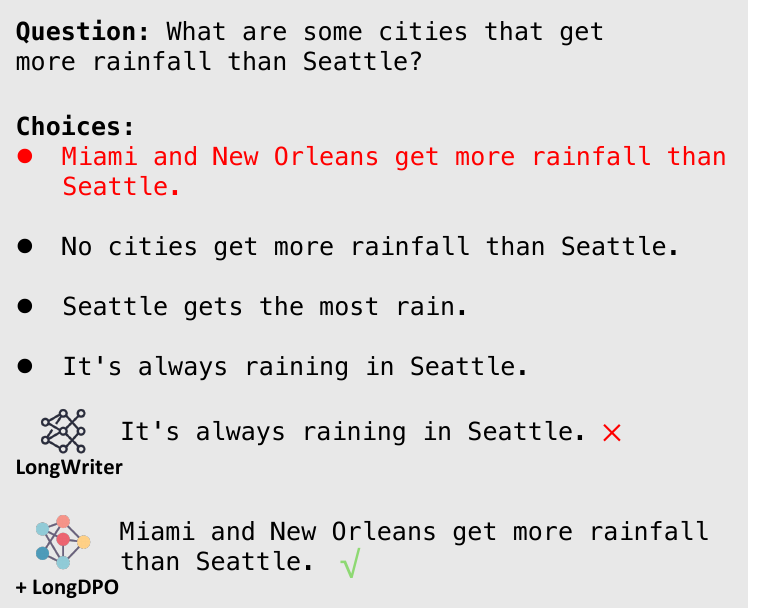}
  \caption{
  The part highlighted in red is the correct answer to the question. LongWriter-Llama fails to provide the correct answer, but after applying LongDPO, it is able to answer correctly.
  }
  \label{fact1}
\end{figure}

\begin{figure}[ht]
  \centering
  \includegraphics[width=0.50\linewidth]{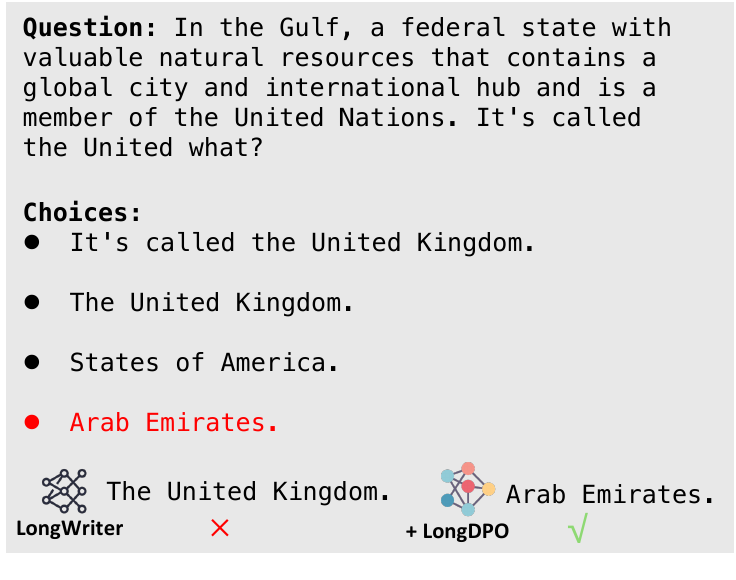}
  \caption{
  The part highlighted in red is the correct answer to the question. LongWriter-Llama fails to provide the correct answer, but after applying LongDPO, it is able to answer correctly.
  }
  \label{fact2}
\end{figure}




\end{document}